\documentclass{article}

\usepackage{arxiv}

\usepackage[utf8]{inputenc} 
\usepackage[T1]{fontenc}    
\usepackage{hyperref}       
\usepackage{url}            
\usepackage{booktabs}       
\usepackage{amsfonts}       
\usepackage{nicefrac}       
\usepackage{microtype}      
\usepackage{lipsum}

\usepackage{bm}
\usepackage{amsmath}
\usepackage{amssymb}
\usepackage{amsfonts}
\usepackage[pdftex]{graphicx}
\def\vec#1{\mathbf{#1}}
\usepackage{algorithm}
\usepackage{algorithmic}

\title{Meta-learning representations for clustering with infinite Gaussian mixture models}

\author{Tomoharu Iwata\\ 
  NTT Communication Science Laboratories, Kyoto, Japan}
\date{}

\begin{document}
\maketitle

\begin{abstract}
For better clustering performance, appropriate representations are critical. Although many neural network-based metric learning methods have been proposed, they do not directly train neural networks to improve clustering performance. We propose a meta-learning method that train neural networks for obtaining representations such that clustering performance improves when the representations are clustered by the variational Bayesian (VB) inference with an infinite Gaussian mixture model. The proposed method can cluster unseen unlabeled data using knowledge meta-learned with labeled data that are different from the unlabeled data. For the objective function, we propose a continuous approximation of the adjusted Rand index (ARI), by which we can evaluate the clustering performance from soft clustering assignments. Since the approximated ARI and the VB inference procedure are differentiable, we can backpropagate the objective function through the VB inference procedure to train the neural networks. With experiments using text and image data sets, we demonstrate that our proposed method has a higher adjusted Rand index than existing methods do.
\end{abstract}

\section{Introduction}

Clustering is an important machine learning task,
in which instances are organized into groups
such that instances in each group are more similar to each other than
to those in other groups.
Clustering has been used in a wide variety of fields~\cite{xu2005survey,berkhin2006survey},
including
natural language processing~\cite{karypis2000comparison,xu2003document},
computer vision~\cite{coleman1979image,chang2017deep},
sensor networks~\cite{abbasi2007survey}, and
marketing~\cite{russell1999fuzzy}.

To improve clustering performance, appropriate representations for the given data are critical.
For example, the Euclidean distance between images based on the original RGB representations
is different from the dissimilarity between images that humans recognize.
To find good representations, many metric learning methods have been proposed.
From recent advances in deep learning, metric learning performance has been improved~\cite{hadsell2006dimensionality,wang2014learning,hoffer2015deep,kumagai2019transfer},
in which neural networks are used for mapping instances from the original space to the representation space.
These methods train neural networks such that
instances with the same labels are located closely in the representation space,
while those with different labels are located further apart.
After representations are obtained with the trained neural networks,
a clustering method is applied to group the instances as post-processing.
Since the existing methods separately perform metric learning and clustering,
the obtained representations might not be suitable for the clustering method.

We propose a meta-learning method for clustering,
in which neural networks for obtaining representations
are trained to directly improve the expected clustering performance.
In the training phase, we use labeled data obtained from various categories.
In the test phase, we are given unlabeled data obtained from unknown categories that are different from the training categories.
Our aim is to improve the clustering performance in the test phase.

\begin{figure*}[t!]
  \centering
  \includegraphics[width=33em]{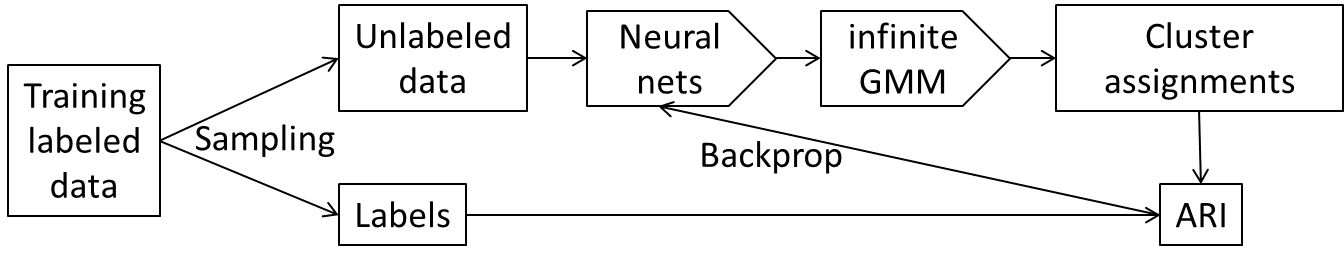}
  \caption{Our training framework. For each epoch, unlabeled data and their corresponding labels are randomly sampled from training labeled data. Cluster assignments of the unlabeled data are estimated by neural networks and the VB inference of an infinite GMM. The continuous approximation of the ARI is calculated using the estimated cluster assignments and the labels. The parameters of the neural networks are updated by backpropagating the ARI.}
  \label{fig:train}
\end{figure*}

Our model obtains the representations of the given unlabeled data using neural networks,
and finds clusters by fitting an infinite Gaussian mixture model (GMM)~\cite{rasmussen2000infinite}
with the obtained representations
based on the variational Bayesian (VB) inference.
With the infinite GMMs, we can automatically estimate the number of clusters from the given data.
The VB inference procedure is the same with prior work~\cite{blei2006variational}.
We newly use the VB inference procedure as a layer in a neural network,
and backpropagate a loss through it to train the neural network.

We evaluate the clustering performance using
the adjusted Rand index (ARI)~\cite{rand1971objective,hubert1985comparing,vinh2010information},
which is a common evaluation measurement for clustering.
We propose a continuous approximation of the ARI to use as the loss.
The neural networks are trained based on a stochastic gradient method
by maximizing the continuous ARI
using an episodic training framework~\cite{finn2017model},
where the test phase is simulated by randomly sampling
a subset of training data for each epoch.
Figure~\ref{fig:train} shows our training framework.

The following are the main contributions of this paper:
\begin{enumerate}
\item We propose a meta-learning method for obtaining representations that directly improves the clustering performance.
\item We newly backpropagate a loss through the VB inference procedure to train neural networks.
\item We propose a continuous approximation of the ARI.
\item We experimentally confirm that our proposed method has better clustering performance than existing methods do.
\end{enumerate}

\section{Related work}

Many unsupervised methods to learn representations
for clustering have been proposed~\cite{iwata2013warped,johnson2016composing,jiang2016variational,yang2019deep,yang2019deep2,min2018survey,nalisnick2016stick,caron2018deep},
in which nonlinear encoders, such as neural networks and Gaussian processes,
and clustering models, such as Gaussian mixture models and k-means,
are combined.
To obtain representations from complex data, flexible encoders like neural networks are needed.
However, when flexible encoders are used,
representations can be modeled even with a single Gaussian distribution,
as in variational autoencoders~\cite{kingma2014autoencoding},
resulting in poor clustering performance.
Therefore, it is difficult to learn flexible encoders
from unlabeled data for clustering.
By contrast, our proposed method can learn flexible encoders
using labeled data by maximizing the expected clustering performance
that is evaluated with label information.

Variational autoencoders~\cite{kingma2014autoencoding}
train neural networks
by maximizing the marginal likelihood based on the VB inference.
By contrast, our proposed method trains neural networks
by maximizing the ARI,
in which the VB inference is used in the forwarding process.
The VB inference steps can be seen as layers
in our neural network-based model
that outputs soft cluster assignments by taking unlabeled data as input.
Our approach gives us a new way of combining the VB inference and deep learning,
and it can be used for improving existing probabilistic models
based on the VB inference in the literature.

Many meta-learning methods have been proposed,
such as model-agnostic meta-learning~\cite{finn2017model}
and matching networks~\cite{vinyals2016matching}.
However, these methods are for supervised learning.
Prototypical network~\cite{snell2017prototypical},
which is a supervised meta-learning method,
tends to cluster instances
in the representation space
according to their categories~\cite{goldblum2020unraveling}.
However, they do not directly train neural networks when clustered.
Although some meta-learning methods for clustering~\cite{hsu2018unsupervised,jiang2019meta,kim2019meta,metz2018meta}
have been proposed,
they are not based on infinite GMMs, and cannot automatically determine the number of clusters.
Although infinite mixture prototypes~\cite{allen2019infinite} are based on infinite GMMs,
since they use an infinite GMM for modeling each category,
they cannot be used for clustering.
Zero-shot learning~\cite{romera2015embarrassingly,xian2017zero,xian2018zero,wang2019survey}
classifies instances belonging to the categories that have no labeled instances.
Our problem setting is different from zero-shot learning in two ways. First, zero-shot learning requires attributes of categories or class-class similarity. Second, the number of categories is known in zero-shot learning.

\section{Proposed method}
\label{sec:proposed}

\subsection{Problem formulation}
\label{sec:problem}

In the training phase, we use labeled data $\mathcal{D}=\{\vec{D}_{t}\}_{t=1}^{T}$ in $T$ tasks,
where $\vec{D}_{t}=\{(\vec{x}_{tn},y_{tn})\}_{n=1}^{N_{t}}$
is the labeled data in the $t$th task,
$\vec{x}_{tn}$ is the $n$th instance, and $y_{tn}\in\{1,\cdots,K_{t}\}$ is its label.
The number of categories $K_{t}$ can be different across tasks.
In the test phase,
we are given unlabeled data $\vec{X}_{*}=\{\vec{x}_{*n}\}_{n=1}^{N_{*}}$,
which are obtained from unknown categories that are different from the training categories.
We do not know the number of clusters in the unlabeled data.
Our aim is to improve the clustering performance of the given unlabeled data in the test phase.

\subsection{Model}

\begin{figure}[t!]
  \centering
  \includegraphics[width=20em]{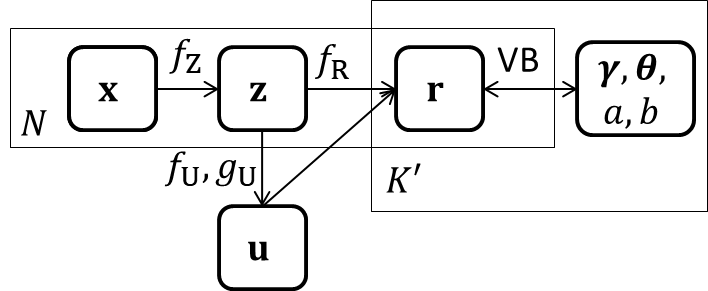}
  \caption{Our model. Soft rectangles are variables. Rectangles indicate repetitions, in which the number at the bottom left corner is the number of repetitions. Our model takes unlabeled data $\{\vec{x}_{n}\}_{n=1}^{N}$ as input. First, the unlabeled data are transformed to $\{\vec{z}_{n}\}_{n=1}^{N}$ in the representation space. Second, task representation $\vec{u}$ is calculated. Third, initial cluster assignments are estimated from instance representations $\{\vec{z}_{n}\}_{n=1}^{N}$ and task representation $\vec{u}$. Fourth, by the VB inference algorithm, cluster assignments $\{\vec{r}_{n}\}_{n=1}^{N}$ and variational parameters $\{\gamma_{k1},\gamma_{k2},\bm{\theta}_{k},a_{k},b_{k}\}_{k=1}^{K'}$ are alternately updated. Note that this figure does not show the generative process,
  but instead shows the forwarding procedure of our model.}
  \label{fig:model}
\end{figure}

Our model takes unlabeled data $\vec{X}=\{\vec{x}_{n}\}_{n=1}^{N}$ as input,
and outputs soft cluster assignments $\vec{R}=\{\vec{r}_{n}\}_{n=1}^{N}$,
where $\vec{r}_{n}=\{r_{nk}\}_{k=1}^{K'}$,
$r_{nk}$ is the probability that the $n$th instance is assigned to the $k$th cluster,
$r_{nk}\geq 0$, $\sum_{k=1}^{K'}r_{nk}=1$,
and $K'$ is the maximum number of clusters.
We omit the task index for simplicity in this subsection.

First, we map each instance to a representation space by an encoder network:
\begin{align}
  \vec{z}_{n}=f_{\mathrm{Z}}(\vec{x}_{n}),
  \label{eq:z}
\end{align}
where $\vec{z}_{n}\in\mathbb{R}^{S}$ is the representation vector of the $n$th instance,
$f_{\mathrm{Z}}$ is a neural network,
and $S$ is the dimension of the representation space.
Neural network $f_{\mathrm{Z}}$ is shared across all tasks.
For neural network $f_{\mathrm{Z}}$, we can use a convolutional neural network
when the given data are images,
and we can use a feed-forward neural network when they are vectors.

Second, we calculate task representation $\vec{u}$, which contains information about all given unlabeled data $\vec{X}$,
by a permutation invariant neural network~\cite{zaheer2017deep}:
\begin{align}
  \vec{u}=g_{\mathrm{U}}\left(\frac{1}{N}\sum_{n=1}^{N}f_{\mathrm{U}}(\vec{z}_{n})\right),
  \label{eq:u}
\end{align}
where $g_{\mathrm{U}}$ and $f_{\mathrm{U}}$ are feed-forward neural networks shared across all tasks.
We use a permutation invariant neural network
since the task representation should not depend on the order of the instances,
and it can handle different numbers of instances $N$.

Third, the initial cluster assignments are calculated using instance representation $\vec{z}_{n}$ and
task representation $\vec{u}$ by a neural network:
\begin{align}
  \log \vec{r}_{n} \propto f_{\mathrm{R}}([\vec{z}_{n},\vec{u}]),
  \label{eq:init_r}
\end{align}
where
$f_{\mathrm{R}}$ is a feed-forward neural network shared across all tasks,
and $[\cdot,\cdot]$ is a vector concatenation.
By concatenating the instance and task representations,
we can obtain initial cluster assignments that consider the relationship between the instance and all given unlabeled data $\vec{X}$.

Fourth, we update cluster assignments by fitting an infinite GMM based on the VB inference.
We assume a spherical infinite GMM
with the following generative process: 
\begin{enumerate}
\item For each cluster $k=1,\cdots,\infty$
  \begin{enumerate}
  \item Draw stick proportion $\eta_{k}\sim\mathrm{Beta}(1,\alpha)$
  \item Set mixture weight $\pi_{k}=\eta_{k}\prod_{k'=1}^{k-1}(1-\eta_{k'})$
  \item Draw mean $\bm{\mu}_{k}\sim\mathcal{N}(\vec{0},\vec{I})$
  \item Draw precision $\beta_{k}\sim\mathrm{Gamma}(1,1)$
  \end{enumerate}
\item For each instance $n=1,\cdots,N$
  \begin{enumerate}
  \item Draw cluster assignment\\ $v_{n}\sim \mathrm{Categorical}(\bm{\pi})$
  \item Draw instance representation\\ $\vec{z}_{n}=\mathcal{N}(\bm{\mu}_{v_{n}},\beta_{v_{n}}^{-1}\vec{I})$
  \end{enumerate}
\end{enumerate}
where
$\mathrm{Beta}$ is the beta distrubtion,
$\mathcal{N}(\bm{\mu},\bm{\Sigma})$ is the Gaussian distribution with mean $\bm{\mu}$ and covariance $\bm{\Sigma}$,
$\mathrm{Gamma}$ is the gamma distribution,
$\mathrm{Categorical}$ is the categorical distribution, and
$\bm{\pi}=\{\pi_{k}\}_{k=1}^{\infty}$.
The mixture weights
are constructed by 
a Dirichlet process prior
with concentration parameter $\alpha$
by a stick breaking process~\cite{sethuraman1994constructive}.
Although we explain our proposed method with spherical Gaussian components
for simplicity and computational efficiency,
our proposed method can also use full covariance Gaussian components.
With the infinite GMM,
the likelihood is given by:
\begin{align}
  p(\vec{X})=\int\int\int p(\bm{\eta})p(\vec{M})p(\bm{\beta})
  \times\prod_{n=1}^{N}\left(\sum_{k=1}^{\infty}p(k|\bm{\eta})p(\vec{x}_{n}|\bm{\mu}_{k},\beta_{k}^{-1}\vec{I})\right)
  d\bm{\eta}d\vec{M}d\bm{\beta},
\end{align}
where
$\bm{\eta}=\{\eta_{k}\}_{k=1}^{\infty}$,
$\vec{M}=\{\bm{\mu}_{k}\}_{k=1}^{\infty}$,
and $\bm{\beta}=\{\beta_{k}\}_{k=1}^{\infty}$.
We assume the following variational posterior distributions
for stick breaking parameter $\eta_{k}$,
mean $\bm{\mu}_{k}$,
precision $\bm{\beta}_{k}$,
and cluster assignment $v_{n}$:
\begin{align}
  q(\eta_{k})=\mathrm{Beta}(\gamma_{k1},\gamma_{k2}),\quad
  q(\bm{\mu}_{k})=\mathcal{N}(\bm{\theta}_{k},\vec{I}),\quad
  q(\beta_{k})=\mathrm{Gamma}(a_{k},b_{k}),\quad
  q(v_{n}=k)=r_{nk}.
\end{align}
Using Jensen's inequality, the evidence lower bound is given by:
\begin{align}
  \log p(\vec{X})
  &\geq\int\int\int q(\bm{\eta})q(\vec{M})q(\bm{\beta})
  \Biggl(
  \sum_{n=1}^{N}
  \Bigl(\sum_{k=1}^{\infty}q(v_{n}=k)
  \left(\log p(k|\bm{\eta})p(\vec{x}_{n}|\bm{\mu}_{k},\beta_{k}^{-1}\vec{I})-\log q(v_{n}=k)\right)
  \Bigr)
  \nonumber\\
  &-\log q(\bm{\eta})q(\vec{M})q(\bm{\beta})\Biggr)
  d\bm{\eta}d\vec{M}d\bm{\beta}.
\end{align}
The update rules of parameters in the variational posterior distributions
are calculated in the closed form by maximizing the evidence lower bound
as follows:
\begin{align}
  \gamma_{k1}&=1+\sum_{n=1}^{N}r_{nk},\quad
  \gamma_{k2}=\alpha+\sum_{n=1}^{N}\sum_{k'=k+1}^{K'}r_{nk'},\quad
  \bm{\theta}_{k} = \frac{\frac{b_{k}}{a_{k}}\sum_{n=1}^{N}r_{nk}\vec{x}_{n}}
     {1+\frac{b_{k}}{a_{k}}\sum_{n=1}^{N}r_{nk}},\nonumber\\
     a_{k}&=1+\frac{S}{2}\sum_{n=1}^{N}r_{nk},\quad
  b_{k}=1+\frac{1}{2}\sum_{n=1}^{N}r_{nk}(\parallel\vec{x}_{n}-\bm{\theta}_{k}\parallel^{2}+S),
 \label{eq:mstep}
\end{align}
\begin{align}
  \log r_{nk}&\propto
  \Psi(\gamma_{k1})-\Psi(\gamma_{k1}+\gamma_{k2})
  -\frac{S}{2}(\Psi(a_{k})-\log(b_{k}))
  -\frac{a_{k}}{2b_{k}}(\parallel\vec{x}_{n}-\bm{\theta}_{k}\parallel^{2}+S)
  \nonumber\\
  &+\sum_{k'=k+1}^{K'}(\Psi(\gamma_{k2})-\Psi(\gamma_{k1}+\gamma_{k2})),
  \label{eq:r}  
\end{align}
where $\Psi$ is the digamma function.
We truncate the number of clusters at $K'$ as in~\cite{blei2004variational}.
The truncated Dirichlet process is shown to closely approximate a true
Dirichlet process for $K'$ that is
large enough relative to the number of instances~\cite{ishwaran2001gibbs}.

Algorithm~\ref{alg:model} shows the forwarding procedures of our model.
We initialize variational parameters by $\vec{a}=1$ and $\vec{b}=1$,
and use Dirichlet process concentration parameter $\alpha=1$ in our experiments.
Since the update rules with the VB inference in
Eqs.~(\ref{eq:mstep},\ref{eq:r})
are differentiable,
we can backpropagate the loss through the VB inference
to train the neural networks.
Since the neural networks in our model are shared across tasks,
our model can handle data in unseen tasks.

\begin{algorithm}[t!]
  \caption{Our model.}
  \label{alg:model}
  \begin{algorithmic}[1]
    \renewcommand{\algorithmicrequire}{\textbf{Input:}}
    \renewcommand{\algorithmicensure}{\textbf{Output:}}
    \REQUIRE{Unlabeled data $\vec{X}$, and maximum number of clusters $K'$}
    \ENSURE{Soft cluster assignments $\vec{R}$}
    \STATE Calculate instance representation $\vec{z}_{n}$ by Eq.~(\ref{eq:z}) for $n:=1$ to $N$.
    \STATE Calculate task representation $\vec{u}$ by Eq.~(\ref{eq:u}).
    \STATE Initialize soft cluster assignments $\vec{r}_{n}$ by Eq.~(\ref{eq:init_r}) for $n:=1$ to $N$.
    \WHILE{End condition is satisfied}
    \STATE Update parameters $\gamma_{k1}, \gamma_{k2}, \bm{\theta}_{k}, a_{k}, b_{k}$
    by Eqs.~(\ref{eq:mstep}) for $k:=1$ to $K'$.
    \STATE Update soft cluster assignments $r_{nk}$ by Eq.~(\ref{eq:r}) for $n:=1$ to $N$ and for $k:=1$ to $K'$.
    \ENDWHILE
  \end{algorithmic}
\end{algorithm}

\subsection{Training}

We train neural networks $f_{\mathrm{Z}}, g_{\mathrm{U}}, f_{\mathrm{U}}, f_{\mathrm{R}}$ in our model
such that the expected test clustering performance is improved.

For the evaluation measurement of the clustering performance,
we use the ARI~\cite{rand1971objective,hubert1985comparing,vinh2010information},
which is a well-known and widely-used evaluation measurement for clustering.
The ARI measures the agreement between the true and estimated cluster assignments,
and it can be used even with different numbers of clusters.
Let $N_{1}$ be the number of pairs of instances that are in different clusters with both the true and estimated assignments,
$N_{2}$ be the number of pairs that are in different clusters in the true assignments but in the same cluster with the estimated assignments,
$N_{3}$ be the number of pairs that are in the same cluster with the true assignments but in different clusters with the estimated assignments,
$N_{4}$ be the number of pairs that are in the same cluster with both the true and estimated assignments.
The ARI is calculated by:
\begin{align}
  \mathrm{ARI}=\frac{2(N_{1}N_{4}-N_{2}N_{3})}{(N_{1}+N_{2})(N_{3}+N_{4})+(N_{1}+N_{3})(N_{2}+N_{4})}.
  \label{eq:ari}
\end{align}
The ARI assumes hard cluster assignments, and is not continuous.
We propose a continuous approximation of the ARI, which
can handle soft assignments.
Let $d_{nn'}$
be the total variation distance~\cite{gibbs2002choosing} between $\vec{r}_{n}$ and $\vec{r}_{n'}$:
\begin{align}
  d_{nn'}=\frac{1}{2}\sum_{k=1}^{K'}|r_{nk}-r_{n'k}|,
  \label{eq:d}
\end{align}
where $0\leq d_{nn'} \leq 1$.
With the continuous approximation of the ARI,
we approximate $N_{1}, N_{2}, N_{3}, N_{4}$ using the total variation distance as follows:
\begin{align}
  \tilde{N}_{1}&=\sum_{n=1}^{N}\sum_{n'=n+1}^{N}I(y_{n}\neq y_{n'})d_{nn'},\\
  \tilde{N}_{2}&=\sum_{n=1}^{N}\sum_{n'=n+1}^{N}I(y_{n}\neq y_{n'})(1-d_{nn'}),\\
  \tilde{N}_{3}&=\sum_{n=1}^{N}\sum_{n'=n+1}^{N}I(y_{n}=y_{n'})d_{nn'},\\
  \tilde{N}_{4}&=\sum_{n=1}^{N}\sum_{n'=n+1}^{N}I(y_{n}=y_{n'})(1-d_{nn'}),
\end{align}
where $y_{n}$ is the true cluster assignment of the $n$th instance.
Let $\bar{\vec{r}}_{n}$ be $\vec{r}_{n}$ with the hard assignment,
i.e., $\bar{r}_{nk}=1$ if $k=\arg\max_{k'}r_{nk'}$ and $\bar{r}_{nk}=0$ otherwise.
When instances $n$ and $n'$ are assigned into the same cluster, $d_{nn'}=0$,
and
when instances $n$ and $n'$ are assigned into different clusters, $d_{nn'}=1$.
Therefore, $\tilde{N}_{i}=N_{i}$ for $i=\{1,2,3,4\}$ with hard assignments,
and the continuous approximation of the ARI:
\begin{align}
  \widetilde{\mathrm{ARI}}=\frac{2(\tilde{N}_{1}\tilde{N}_{4}-\tilde{N}_{2}\tilde{N}_{3})}{(\tilde{N}_{1}+\tilde{N}_{2})(\tilde{N}_{3}+\tilde{N}_{4})+(\tilde{N}_{1}+\tilde{N}_{3})(\tilde{N}_{2}+\tilde{N}_{4})},
  \label{eq:smoothed_ari}
\end{align}
is a natural extension of the ARI in Eq.~(\ref{eq:ari}).

The objective function to be maximized is the following expected test approximated ARI,
\begin{align}
  \hat{\bm{\Psi}}=\arg\max_{\bm{\Psi}}\mathbb{E}_{(\vec{X},\vec{y})\sim\mathcal{D}}
  [\widetilde{\mathrm{ARI}}(\vec{y},\vec{R}(\vec{X}))],
\end{align}
where $\bm{\Psi}$ is a set of parameters of neural networks in our model,
$\mathbb{E}$ is the expectation,
and $\vec{R}(\vec{X})$ is the soft clustering assignments estimated by our model given $\vec{X}$.
The expectation takes over data $\vec{X}$ and their category labels $\vec{y}$,
which is calculated by randomly sampling data and their labels
from training data $\mathcal{D}$.
Algorithm~\ref{alg:train} shows the training procedures of our model with the episodic training framework.
For each training epoch,
first, we randomly construct a clustering task from the training data in Lines 3--6.
Then, we evaluate the clustering perofrmance of our model on the task
based on the continuous approximation of the ARI in Line 7--8,
and update model parameters to maximize the performance in Line 9.

\begin{algorithm}[t!]
  \caption{Training procedure of our model.}
  \label{alg:train}
  \begin{algorithmic}[1]
    \renewcommand{\algorithmicrequire}{\textbf{Input:}}
    \renewcommand{\algorithmicensure}{\textbf{Output:}}
    \REQUIRE{Training data $\mathcal{D}$, maximum number of clusters $K'$}
    \ENSURE{Trained model parameters $\bm{\Psi}$}
    \STATE Initialize model parameters $\bm{\Psi}$.
    \WHILE{End condition is satisfied}
    \STATE Randomly select a task index $t$ from $\{1,\cdots,T\}$.
    \STATE Randomly choose number of categories $K$ from $\{1,\cdots,\min(K',K_{t})\}$.
    \STATE Randomly select $K$ categories from $\{1,\cdots,K_{t}\}$.
    \STATE Construct unlabeled data $\vec{X}$ and their category labels $\vec{y}$ from the selected categories.
    \STATE Estimate soft cluster assignments $\vec{R}(\vec{X})$ by our model (Algorithm~\ref{alg:model}) using unlabeled data $\vec{X}$. 
    \STATE Calculate loss $-\widetilde{\mathrm{ARI}}(\vec{y},\vec{R}(\vec{X}))$ and its gradient using estimated assignments $\vec{R}(\vec{X})$ and labels $\vec{y}$.\hspace{-1em}
    \STATE Update model parameters $\bm{\Psi}$ using the loss and gradient by a stochastic gradient descent method.
    \ENDWHILE
  \end{algorithmic}
\end{algorithm}



\section{Experiments}
\label{sec:experiments}

\subsection{Data}

We evaluated our proposed method with four datasets:
Patent, Dmoz, Omniglot, and Mini-imagenet.
The Patent dataset consisted of patents published
in Japan from January to March in 2004,
which were categorized by International Patent Classification.
Each patent was represented by bag-of-words.
We omitted words that occurred in fewer than 100 patents,
omitted patents with fewer than 100 unique words, and
omitted categories with fewer than 10 patents.
The number of instances, attributes, and categories were
5477, 3201, and 248, respectively.
The Dmoz dataset consisted of webpages crawled in 2006
from the Open Directory Project~\cite{lorenzetti2019dmoz,lorenzetti2009semi}
\footnote{The data were obtained from \url{https://data.mendeley.com/datasets/9mpgz8z257/1}.}.
Each webpage was categorized in a web directory,
and represented by bag-of-words.
We omitted words that occurred in fewer than 300 webpages,
omitted webpages with fewer than 300 unique words,
and omitted categories with fewer than 10 webpages.
The number of instances, attributes, and categories were
15159, 17659, and 354, respectively.
The Omniglot dataset~\cite{lake2015human} consisted of
hand-written images of 964 characters from 50 alphabets (real and fictional).
There were 20 images for each character category.
Each image was represented by gray-scale with 105 $\times$ 105 pixels.
The number of instances, attributes, and categories were
19280, 11025, and 964, respectively.
The Mini-imagenet dataset
consisted of 101 images from 100 categories subsampled from
the original mini-imagenet data~\cite{vinyals2016matching}.
Each image was represented by RGB with 84 $\times$ 84 pixels.
The number of instances, attributes, and categories were
10100, 21168, and 100, respectively.

For each dataset,
we randomly used 60\% of the categories for training,
20\% for validation,
and the remaining categories for testing.
For each task in the validation and test data,
we first randomly selected the number of categories from two to ten,
and we then randomly selected categories.
We performed ten experiments with different data splits
for each dataset.

\subsection{Comparing methods}

We compared our proposed method
with prototypical networks~\cite{snell2017prototypical} (Proto),
autoencoders (AE),
Siamese networks~\cite{hadsell2006dimensionality} (Siamese),
triplet networks~\cite{wang2014learning,hoffer2015deep} (Triplet),
neural network-based classifiers (NN),
variational deep embedding~\cite{jiang2016variational} (VaDE),
principal component analysis (PCA),
and Fisher linear discriminant analysis (FLDA).
Proto, AE, Siamese, Triplet, and VaDE were neural network-based
representation learning methods,
and PCA and FLDA were linear methods.
We found cluster assignments by performing the VB inference
with an infinite GMM after the representations were obtained by each method except for VaDE.

Proto is a meta-learning method for classification,
in which the encoder network
is trained such that the classification performance improves
when instances are classified by the proximity from category centroids
in a representation space.
Proto tends to cluster instances
in the representation space
according to their categories~\cite{goldblum2020unraveling}.
AE finds representations
by minimizing the reconstruction error.
Siamese is a deep metric learning method,
where the encoder network is trained such that
the distance with the same categories becomes small
while that with different categories becomes large.
Triplet is another deep metric learning method,
in which
the distance with the same categories becomes smaller than
that with different categories.
NN is a neural network-based classifier, where
the output layer is a linear layer with the softmax function,
and the cross-entropy loss is used.
VaDE is an unsupervised neural network-based clustering method
that combines variational autoencoders and GMMs,
where the neural network parameters are optimized by
maximizing the evidence lower bound.
PCA embeds instances such that variance is maximized.
FLDA embeds instances such that the between-class variance
is maximized while the within-class variance is minimized.
AE, VaDE and PCA are unsupervised methods,
in which category label information is not used,
and the others are supervised methods.
Note that the supervised methods, including our proposed method,
use label information in training data, but
do not use label information in test data for clustering.
Proto, AE, Siamese, and Triplet
were trained with the episodic training framework,
and VaDE was trained using unlabeled test data.

\subsection{Setting}

We used three-layered feed-forward neural networks
with 256 hidden units for
$f_{\mathrm{Z}}$, $f_{\mathrm{U}}$, $g_{\mathrm{U}}$,
and $f_{\mathrm{R}}$.
The output layer size of $f_{\mathrm{U}}$ was 256.
For image datasets, we additionally used 
four-layered convolutional neural networks of
filter size 32 and kernel size three
for $f_{\mathrm{Z}}$.
The number of dimensions of the representation space was ten.
For the activation function,
we used rectified linear unit $\mathrm{ReLU}(x)=\max(0,x)$.
We used the same encoder network $f_{\mathrm{Z}}$ for all
neural network-based methods, i.e., Proto, AE, Siamese, Triplet, NN, VaDE
and our proposed method.
The number of VB inference steps was ten.
We optimized using Adam~\cite{kingma2014adam}
with learning rate $10^{-3}$,
and dropout rate $10^{-1}$~\cite{srivastava2014dropout}.
We initialized the encoder network $f_{\mathrm{Z}}$
of our proposed method by using Proto.
The validation data were used for early stopping,
for which the maximum number of training epochs was 1,000.
We set the maximum number of clusters at $K'=10$.
Our implementation was based on PyTorch~\cite{paszke2017automatic}.

\subsection{Results}

Table~\ref{tab:ari_method} shows the test ARI.
Our proposed method had the best ARI with all datasets.
This result indicates that our proposed method
can find clusters using an infinite GMM 
by training the neural networks to maximize the expected test ARI
when clustered by the infinite GMM.
Proto had the second best performance.
Proto calculates the category (cluster) assignments
based on the proximity to their category centroids,
which is the same with the GMM.
Therefore, the instance representations by Proto were
more appropriate than those by Siamese or Triplet,
which does not consider category centroids.
AE, VaDE and PCA did not perform well
since they are unsupervised methods.
The test ARI by FLDA was also low since it is a linear method.

\begin{table*}[t!]
  \centering
  \caption{Test ARI averaged over ten experiments. Values in bold are not statistically significantly different at the 5\% level from the best performing method in each row according to a paired t-test.}
  \label{tab:ari_method}
  \vspace{0.5em}
    {\tabcolsep=0.5em\begin{tabular}{lrrrrrrrrr}
    \hline
    & Ours & Proto & AE & Siamese & Triplet & NN & VaDE & PCA & FLDA \\
    \hline
Patent & {\bf 0.550} & 0.480 & 0.000 & 0.016 & 0.114 & 0.236 & 0.000 & 0.181 & 0.043\\
Dmoz & {\bf 0.469} & 0.313 & 0.000 & 0.000 & 0.018 & 0.161 & 0.001 & 0.062 & 0.028\\
Omniglot & {\bf 0.869} & 0.815 & 0.112 & 0.006 & 0.308 & 0.480 & 0.026 & 0.124 & 0.090\\
Mini-imagent & {\bf 0.112} & 0.052 & 0.034 & 0.000 & 0.013 & 0.064 & 0.000 & 0.005 & 0.014\\
\hline
\end{tabular}}
\end{table*}

\begin{figure*}[t!]
\renewcommand{\arraystretch}{0.1}
  \centering
  {\tabcolsep=-0.5em
    \begin{tabular}{cccc}
      \multicolumn{4}{c}{Task 1} \\
    \includegraphics[width=12.7em]{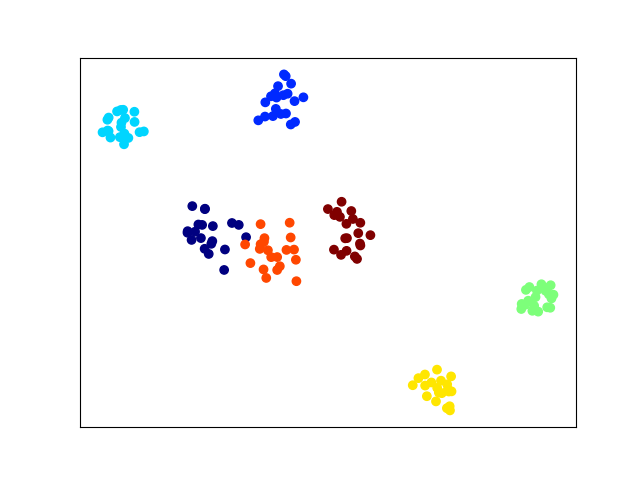}&
    \includegraphics[width=12.7em]{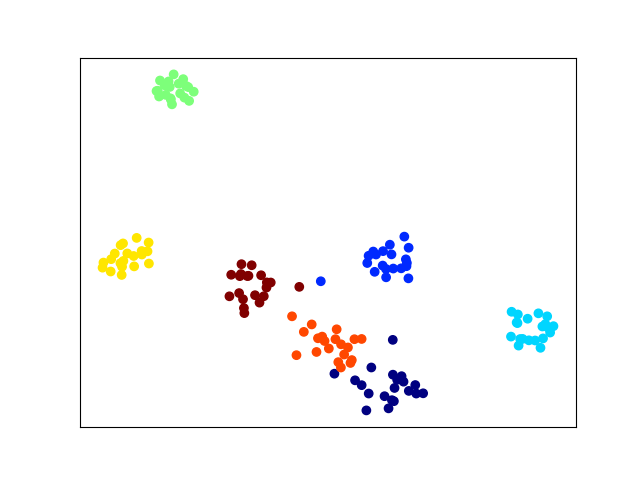}&
    \includegraphics[width=12.7em]{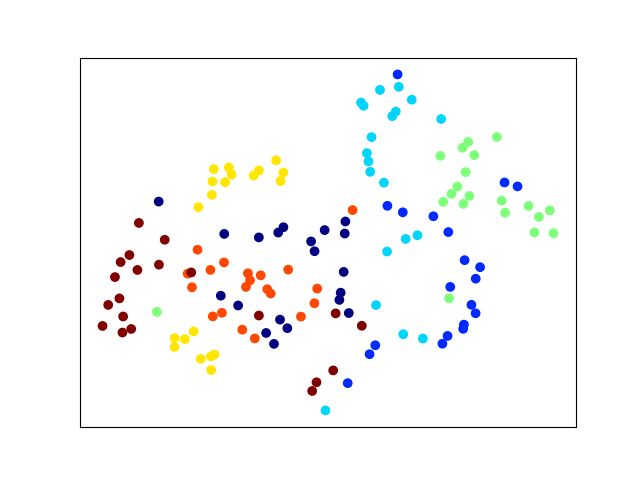}&
    \includegraphics[width=12.7em]{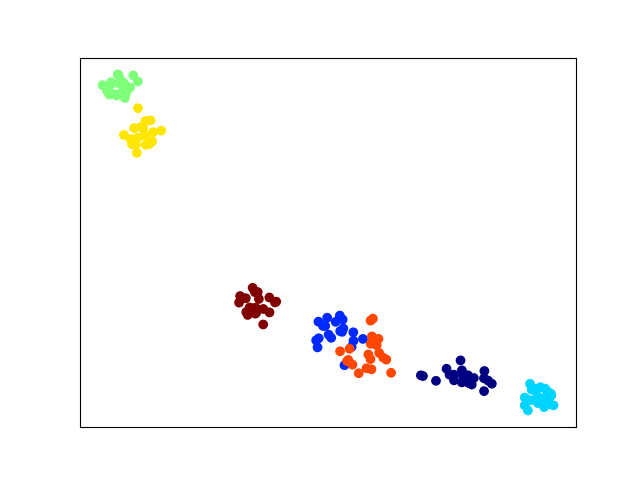}\\
     \multicolumn{4}{c}{Task 2}\\    
    \includegraphics[width=12.7em]{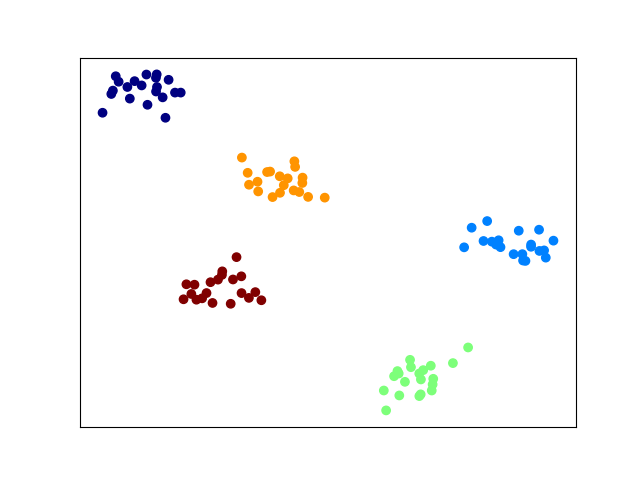}&
    \includegraphics[width=12.7em]{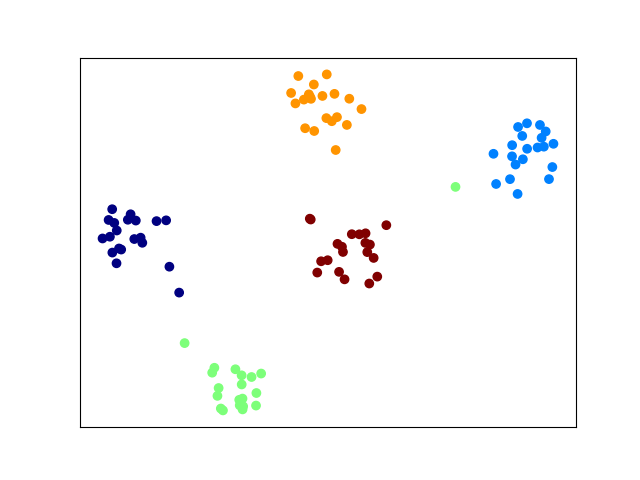}&
    \includegraphics[width=12.7em]{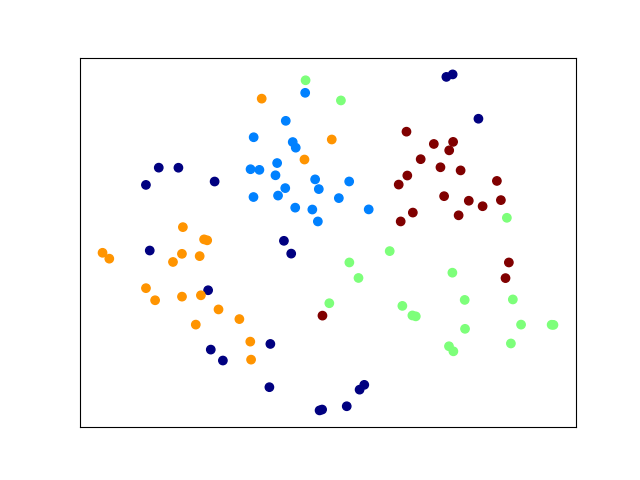}&
    \includegraphics[width=12.7em]{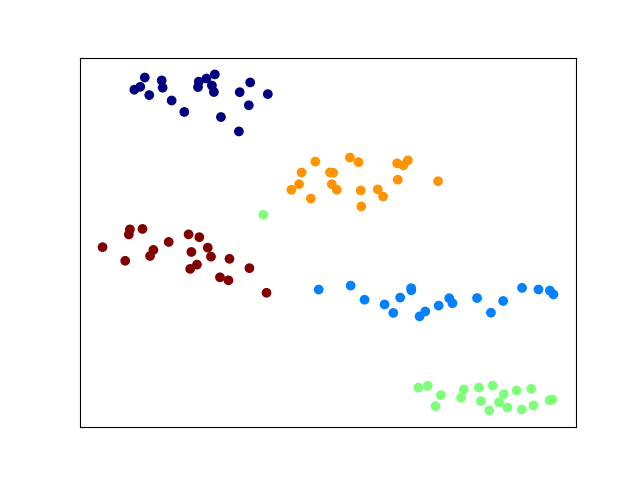}\\
    (a) Ours & (b) Proto & (c) AE & (d) Triplet \\
  \end{tabular}}
  \caption{Two-dimensional visualization of test instances in the representation space by t-SNE~\cite{maaten2008visualizing} with two target tasks in the Omniglot dataset. Each point is an instance, and the color represents its category.}
  \label{fig:vis}
\renewcommand{\arraystretch}{1.0}  
\end{figure*}

Figure~\ref{fig:vis} shows the visualization of test data
in the representation space with the Omniglot dataset.
Our proposed method appropriately encoded the instances
such that data from the same category can be modeled by
a Gaussian distribution, which resulted in better clustering performance
by an infinite GMM.
AE failed to cluster instances since it is an unsupervised method.
With Triplet, instances from the same categories were located closely.
However,
instances from different categories were not well separated,
and the shape of each category was far from a spherical
Gaussian distribution.
Therefore, the clustering performance of Triplet was worse than that of our
proposed method as shown in Table~\ref{tab:ari_method}.

\begin{figure*}[t!]
  \centering
  \begin{tabular}{cccc}
    \includegraphics[width=10.2em]{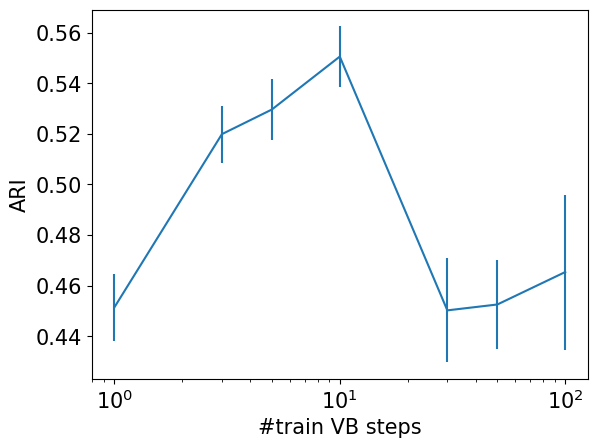}&
    \includegraphics[width=10.2em]{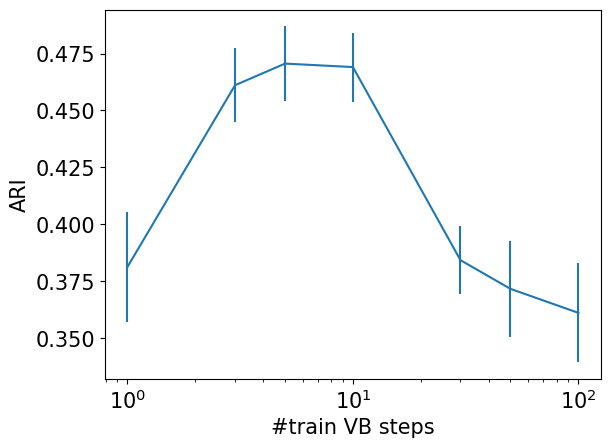}&
    \includegraphics[width=10.2em]{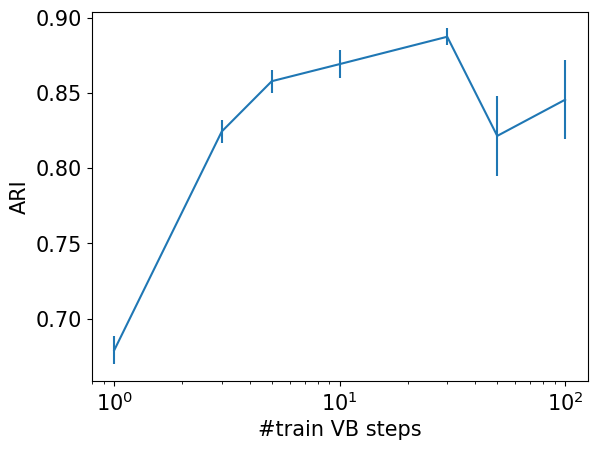}&
    \includegraphics[width=10.2em]{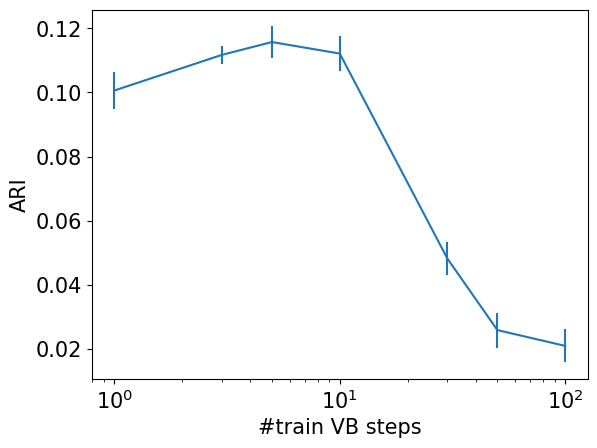}\\
    (a) Patent & (b) Dmoz & (c) Omniglot & (d) Mini-imagenet\\
  \end{tabular}
  \caption{Averaged test ARI with different numbers of VB steps in the training phase. The bar shows the standard error.}
  \label{fig:ari_train_epoch}
\end{figure*}

\begin{figure*}[t!]
  \centering
  \begin{tabular}{cccc}
    \includegraphics[width=10.2em]{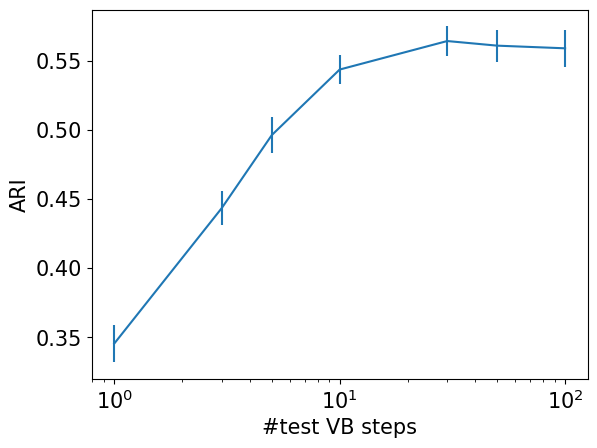}&
    \includegraphics[width=10.2em]{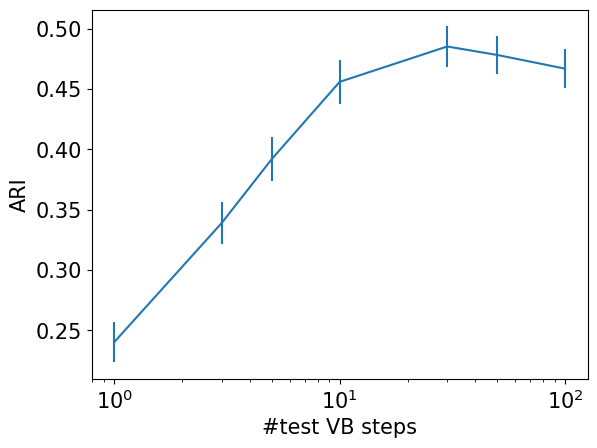}&
    \includegraphics[width=10.2em]{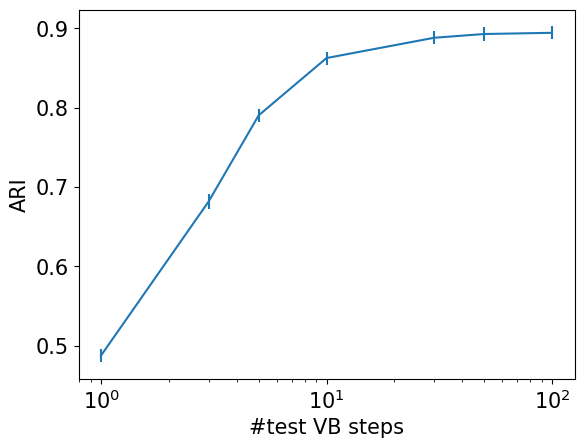}&
    \includegraphics[width=10.2em]{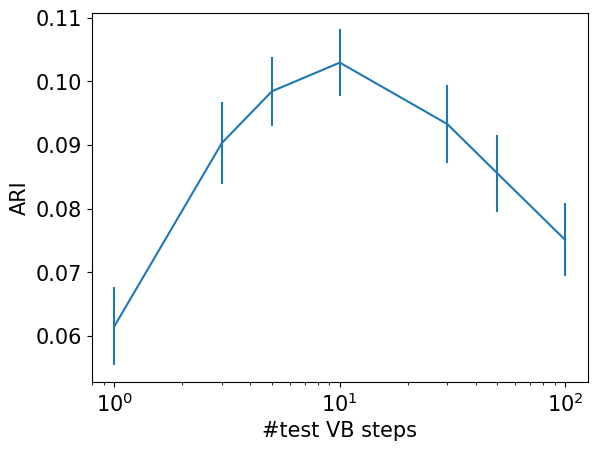}\\
    (a) Patent & (b) Dmoz & (c) Omniglot & (d) Mini-imagenet\\
  \end{tabular}
  \caption{Averaged test ARI and its standard error with different numbers of VB steps in the test phase.}
  \label{fig:ari_test_epoch}
\end{figure*}

\begin{figure*}[t!]
  \centering
  \begin{tabular}{cccc}
    \includegraphics[width=10.2em]{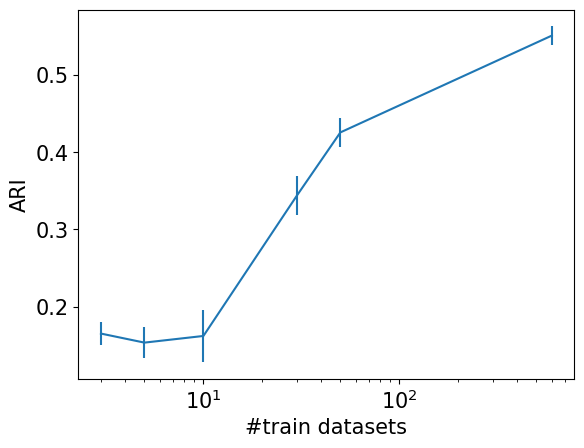}&
    \includegraphics[width=10.2em]{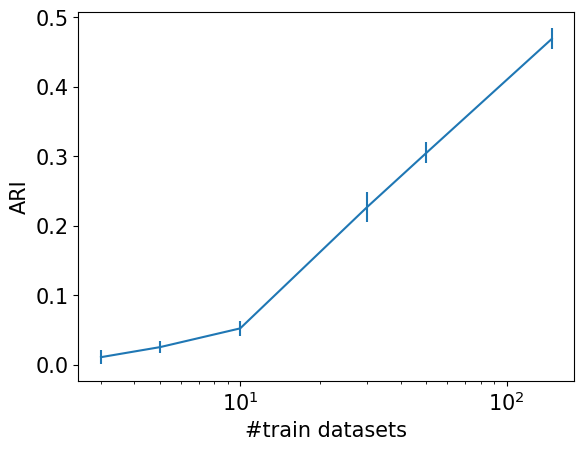}&
    \includegraphics[width=10.2em]{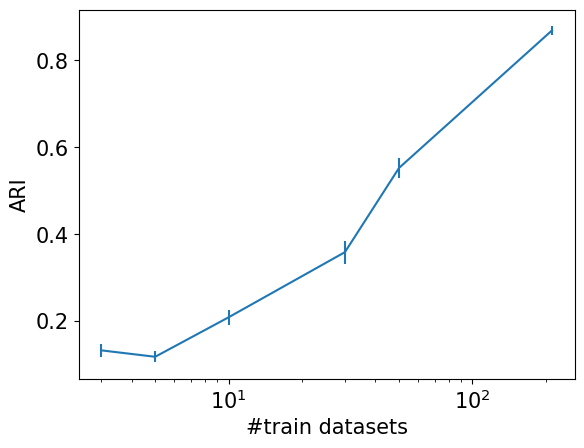}&
    \includegraphics[width=10.2em]{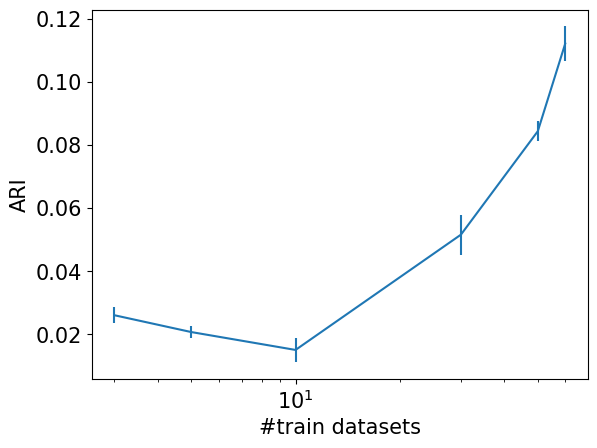}\\
    (a) Patent & (b) Dmoz & (c) Omniglot & (d) Mini-imagenet\\
  \end{tabular}
  \caption{Averaged test ARI and its standard error with different numbers of categories in the training data.}
  \label{fig:ari_n_dataset}
\end{figure*}

Figure~\ref{fig:ari_train_epoch} shows
the test ARI with different numbers of VB steps in the training phase.
The performance was the best with five or ten VB steps.
With a small number of VB steps (one or three),
the infinite GMM could not find clusters.
With a large number of VB steps (50 or 100),
it was difficult to backpropagate the loss to neural networks.
Figure~\ref{fig:ari_test_epoch} shows
the test ARI with different numbers of VB steps in the test phase,
in which we used ten VB steps in the training phase.
The test ARI around ten test VB steps was high.
This result indicates that
the different numbers of VB steps between the training and test phases
can deteriorate the performance.
Figure~\ref{fig:ari_n_dataset} shows
the test ARI with different numbers of categories in the training data.
As the number of categories increased,
the performance increased.
This result implies that
it is important to use data from many categories for training
to improve performance.
  
\begin{table*}[t!]
  \centering
  \caption{Averaged test ARI and its standard error on the ablation study. `Ours' is our proposed method. `w/o $f_{\mathrm{R}}$' is Ours without initial cluster assignment calculation by Eq.~(\ref{eq:r}), in which initial cluster assignments are randomly determined using a standard Gaussian distribution. `w/o Pretrain' is Ours without pretraining by using Proto. `w/ EM' is Ours that infers a GMM with ten clusters by the EM algorithm. `w/ Prob' is Ours that calculates the continuous ARI based on the probability that two instances are clustered in different clusters, $d_{nn'}=1-\sum_{k=1}^{K'}r_{nk}r_{n'k}$, instead of the total variation distance. Values in bold typeface are not statistically significantly different at the 5\% level from the best performing method in each row according to a paired t-test.}
  \label{tab:ablation}
  \vspace{0.5em}
  (a) Ten-dimensional representation space\\
    \begin{tabular}{lrrrrr}
    \hline
    & Ours & w/o $f_{\mathrm{R}}$ & w/o Pretrain & w/ EM & w/ Prob\\
    \hline
Patent & {\bf 0.550$\pm$0.012} & {\bf 0.539$\pm$0.009} & 0.330$\pm$0.016 & 0.472$\pm$0.011 & {\bf 0.546$\pm$0.011}\\
Dmoz & {\bf 0.469$\pm$0.015} & {\bf 0.462$\pm$0.014} & 0.000$\pm$0.000 & 0.411$\pm$0.017 & 0.427$\pm$0.015\\
Omniglot & {\bf 0.869$\pm$0.009} & {\bf 0.835$\pm$0.023} & 0.763$\pm$0.016 & {\bf 0.865$\pm$0.013} & {\bf 0.858$\pm$0.008}\\
Mini-imagenet & {\bf 0.112$\pm$0.005} & {\bf 0.092$\pm$0.008} & 0.076$\pm$0.005 & {\bf 0.068$\pm$0.008} & 0.105$\pm$0.005\\
    \hline
    \end{tabular}
    \\
  \vspace{0.5em}
  (b) Two-dimensional representation space\\
  \begin{tabular}{lrrrrr}
    \hline
    & Ours & w/o $f_{\mathrm{R}}$ & w/o Pretrain & w/ EM & w/ Prob\\
    \hline
    Patent & {\bf 0.391$\pm$0.013} & 0.371$\pm$0.015 & 0.302$\pm$0.011 & 0.348$\pm$0.008 & {\bf 0.389$\pm$0.016}\\
    Dmoz & {\bf 0.290$\pm$0.015} & 0.265$\pm$0.013 & 0.000$\pm$0.000 & 0.261$\pm$0.014 & 0.268$\pm$0.014\\
    Omniglot & {\bf 0.651$\pm$0.011} & 0.582$\pm$0.008 & 0.627$\pm$0.009 & 0.633$\pm$0.013 & 0.618$\pm$0.010\\
    Mini-imagenet &{\bf 0.084$\pm$0.007} & 0.068$\pm$0.007 & {\bf 0.091$\pm$0.010} & 0.063$\pm$0.003 & {\bf 0.078$\pm$0.007}\\
    \hline
  \end{tabular}
\end{table*}

Table~\ref{tab:ablation} shows the test ARI on the ablation study.
When cluster assignments were randomly initialized, the performance was low
especially with the two-dimensional representation space.
This result demonstrates that
it is important to determine appropriate initial assignments
using neural networks,
$f_{\mathrm{U}}$, $g_{\mathrm{U}}$, and $f_{\mathrm{R}}$,
since the backpropagation through many VB steps is difficult.
Without pretraining by using Proto, the test ARI was worse.
Since our model has multiple VB steps without trainable parameters
after neural networks,
there is a risk of getting stuck in bad local optima.
The pretraining mitigated this risk,
and improved performance.
Note that the proposed method's performance with pretraining
is significantly better than ProtoNet as shown in Table~\ref{tab:ari_method}.
Since the EM algorithm of GMMs is also differentiable
as the VB inference,
we can use the EM algorithm for maximizing the likelihood by
fixing the number of clusters with GMMs.
However, the performance with the EM algorithm was worse
because it cannot infer the number of clusters from the given data.
When the continuous ARI is calculated
based on the probability that two instances are clustered in different clusters, $d_{nn'}=1-\sum_{k=1}^{K'}r_{nk}r_{n'k}$, instead of the total variation distance,
the performance was worse.
  
\begin{table}[t!]
  \centering
  \caption{Training computational time in seconds.}
  \label{tab:train_time}
  \vspace{0.5em}
  {\tabcolsep=0.3em\begin{tabular}{lrrrrrrrr}
    \hline
    & Ours & Proto & AE & Siamese & Triplet \\
    \hline
Patent & 6066 & 1542 & 1223 & 1528 & 1591 \\
Dmoz & 13861 & 4939 & 4513 & 4683 & 5107 \\
Omniglot & 23564 & 7891 & 8366 & 7106 & 7627 \\
Mini-imagent & 49983 & 20570 & 20835 & 20384 & 21053 \\
    \hline
  \end{tabular}}
\end{table}

Table~\ref{tab:train_time} 
shows the average computational time for training
with a GTX 1080Ti GPU. Since our proposed method requires VB inference steps
for training, it takes longer to train than other methods do.
The average test computational time by our proposed method was
1.08, 2.12, 2.01, and 14.89 seconds with the Patent, Dmoz, Omniglot, and Mini-imagenet data sets, respectively.
The test computational time was almost the same with other neural network-based methods
since all instances were encoded using neural networks with the same structure.

\section{Conclusion}

We proposed a meta-learning method for obtaining representations for clustering.
Our proposed method trains the encoder neural network such that
the clustering performance improves when the representations are clustered by an infinite Gaussian mixture model.
Experiments on four data sets confirmed that our proposed method had better clustering performance
than existing methods did.
For future work,
we will apply our framework for meta-learning neural networks through the variational Bayesian inference
to other probabilistic models.

\bibliographystyle{abbrv}
\bibliography{icml2021gmm}

\end{document}